\documentclass[10pt]{article}

\usepackage[margin=1in]{geometry}
\usepackage[T1]{fontenc}
\usepackage{lmodern}
\usepackage{microtype}
\usepackage{setspace}
\usepackage{amsmath, amssymb}
\usepackage{booktabs}
\usepackage{siunitx}
\usepackage{graphicx}
\usepackage{subcaption}
\usepackage{float}
\usepackage{placeins} 

\usepackage{tikz}
\usetikzlibrary{arrows.meta, positioning, fit, backgrounds}

\graphicspath{{image/}{images/}{report_artifacts_alllabels_v2/}{report/files/report_artifacts_alllabels_v2/}}

\usepackage{algorithm}
\usepackage{algpseudocode}
\usepackage{listings}
\usepackage{xcolor}
\usepackage[numbers,sort&compress]{natbib}
\usepackage[hidelinks]{hyperref}
\usepackage{url}
\usepackage{xurl}
\normalfont

\title{\textbf{REDNET-ML: A Multi-Sensor Machine Learning Pipeline for Harmful Algal Bloom Risk Detection Along the Omani Coast}}
\author{Ameer Alhashemi \hspace{1em} \\ \small School of Computer Science, University of Birmingham}
\date{March 2026}

\begin{document}
\pagenumbering{gobble}
\maketitle
\vspace{-1.5em}   
\pagenumbering{arabic}

\begin{abstract}
Harmful algal blooms (HABs) can threaten coastal infrastructure, fisheries, and desalination-dependent water supplies. This project (REDNET-ML) develops a reproducible machine learning pipeline for HAB \emph{risk} detection along the Omani coastline using multi-sensor satellite data and non-leaky evaluation. The system fuses (i) Sentinel-2 optical ``chips'' (high spatial resolution) processed into spectral indices and texture signals, (ii) MODIS Level-3 ocean color and thermal indicators, and (iii) learned image evidence from object detectors trained to highlight bloom-like patterns. A compact decision-fusion model (CatBoost) integrates these signals into a calibrated probability of HAB risk, which is then consumed by an end-to-end inference workflow and a risk-field viewer that supports operational exploration by site (plant) and time. The report documents the motivation, related work, methodological choices (including label mining and strict split strategies), implementation details, and a critical evaluation using AUROC/AUPRC, confusion matrices, calibration curves, and drift analyses that quantify distribution shift in recent years.
\end{abstract}

\vspace{0.5em}

\section{Introduction and Motivation}

Coastal Harmful Algal Blooms (HABs) are episodic biological events in which rapid phytoplankton growth alters water quality, depletes oxygen, and can introduce toxins. These events have well-documented ecological and socio-economic consequences, including fisheries disruption, shoreline impacts, and risks to industrial intakes (e.g., desalination and cooling-water systems) that depend on reliable coastal water quality \citep{anderson_hab_impacts}. Along the Omani coastline and the broader Gulf of Oman, HAB events are a recurring regional concern and motivate the need for systematic monitoring and decision support.

Satellite remote sensing offers a scalable approach: ocean color products capture chlorophyll and related optical proxies over large areas, while higher-resolution imagery can resolve nearshore structure and localized features that may be relevant to plant-level risk \citep{esa_s2_guide,nasa_oc_modis_l3}. However, translating raw imagery into an actionable signal remains non-trivial: (i) HAB ground truth is sparse and often delayed, (ii) labels can be noisy or only available at coarse temporal resolution, and (iii) na\"ive train/test splits can severely overestimate performance due to temporal and spatial leakage (e.g., learning scene-specific artifacts rather than the underlying phenomenon).

REDNET-ML addresses this by building an \emph{end-to-end, non-leaky, and inspectable} pipeline that outputs a plant-centric HAB risk probability. The key ideas are: multi-sensor feature engineering (Sentinel indices fused with MODIS ocean color/thermal predictors) \citep{esa_s2_guide,nasa_oc_modis_l3,mcf_ndwi_1996,hu_fai_2009}; image evidence as detector scores (detectors act as evidence generators rather than final labelers) \citep{ren_fasterrcnn_2015,liu_ssd_2016,ultralytics_yolov8}; decision-level fusion with strict splitting (CatBoost with group-safe and temporal splits) \citep{prokhorenkova_catboost_2018}; and operationalization via an end-to-end inference workflow and a risk-field viewer for inspection. Each stage is implemented as reproducible notebooks and scripts, with exported evaluation artifacts (ROC/PR, confusion, calibration, feature importance, SHAP) enabling transparent examination.

\section{Project Background and Context}

\subsection{Remote sensing for HAB monitoring}
HAB detection from remote sensing typically relies on two classes of signals. First, \emph{ocean color} products approximate phytoplankton biomass and optical properties (e.g., chlorophyll-\textit{a}, diffuse attenuation, fluorescence proxies). MODIS Level-3 products provide consistent gridded summaries suitable for time-series monitoring at regional scales \citep{nasa_oc_modis_l3}. Second, \emph{high-resolution multispectral imagery} (e.g., Sentinel-2) represents nearshore variability at 10--20\,m resolution and enables local gradients, indices, and spatial structure signals that may matter for plant-level risk \citep{esa_s2_guide}.

In practice, operational monitoring often mixes remote sensing with in-situ sampling, expert review, and incident reports. This creates a common ML constraint: the phenomenon is continuous, but labels are sparse, delayed, and heterogeneous. Under this regime, the primary failure modes are overfitting and false confidence driven by data dependence.

This project also sits alongside a broader body of ML-driven HAB work in the region and in operational settings, including forecasting-oriented studies in the Persian Gulf / Gulf of Oman \citep{hab_ai_persian_gulf_gulf_oman}, physics-aware prediction efforts for the Middle East \citep{middle_east_hab_submesoscale}, and end-to-end systems that integrate remote sensing with ML for detection and monitoring \citep{habnet_system,harmful_algal_information_system}. Open implementations for forecasting related bio-optical indicators (e.g., chlorophyll-\textit{a}) also motivate reproducibility expectations \citep{djawadi_chla_github}.

\subsection{Spectral indices and contextual predictors}
Index-based features remain valuable because they encode domain structure with minimal model complexity. The Normalized Difference Water Index (NDWI) helps separate water from non-water reflectance regimes and is widely used in surface-water delineation \citep{mcf_ndwi_1996}. For algae-related signals, indices based on green/red/NIR contrast can act as weak indicators under appropriate conditions; the Floating Algae Index (FAI) is one example using a baseline-corrected NIR anomaly \citep{hu_fai_2009}. In REDNET-ML, indices serve two roles: explainable fusion features and support for label mining heuristics when trusted labels are limited.

Beyond indices, contextual predictors (sea surface temperature, seasonal encoding, and ocean color proxies) improve separability and robustness. Seasonality is encoded as sine/cosine of month to learn recurring annual structure without using absolute timestamps.

\subsection{Learning from images: detectors as evidence generators}
Many HAB approaches treat each patch as a binary classification problem. REDNET-ML uses object detectors differently: detectors act as \emph{evidence generators} that convert imagery into a small set of robust score summaries reflecting bloom-like spatial structure. This is advantageous under label scarcity because it avoids learning an end-to-end deep classifier from limited positives while still capturing image-driven cues. Detector families used include Faster R-CNN and SSD/YOLO-style one-stage models \citep{ren_fasterrcnn_2015,liu_ssd_2016,ultralytics_yolov8}; diversity reduces reliance on a single inductive bias.

\subsection{Evaluation pitfalls: leakage and temporal shift}
Random splits often overestimate performance in time/space-correlated remote sensing data. Leakage can occur if chips from the same overpass appear in both train and test, or if derived features are computed in a way that uses future information. REDNET-ML addresses this with \emph{group-safe CV} (scene-aware folds) and \emph{time-based CV} (train on earlier periods, test on later periods) to approximate forward deployment. The project also reports drift (PSI and KS distance) to contextualize threshold stability under year-to-year distribution shift.

\section{Methodology and Specifications}

\subsection{System specification}
The system maps $(\text{plant}, t)$ to a calibrated risk probability $\hat{p}_{t}$ and a discrete alert state:
\begin{align}
\hat{p}_{t} &= f_{\theta}\big(\mathbf{x}^{\text{S2}}_{t}, \mathbf{x}^{\text{MODIS}}_{t}, \mathbf{s}^{\text{det}}_{t}\big) \in [0,1], \\
\text{state}(\hat{p}_{t}) &=
\begin{cases}
\textsf{NORMAL} & \hat{p}_{t} < \tau_{\text{watch}} \\
\textsf{WATCH}  & \tau_{\text{watch}} \le \hat{p}_{t} < \tau_{\text{action}} \\
\textsf{ACTION} & \hat{p}_{t} \ge \tau_{\text{action}} \\
\end{cases}
\end{align}
where $\mathbf{x}^{\text{S2}}_{t}$ are Sentinel-derived features, $\mathbf{x}^{\text{MODIS}}_{t}$ are Level-3 context predictors, and $\mathbf{s}^{\text{det}}_{t}$ are detector score summaries. Thresholds are selected from validation PR curves under a minimum-recall constraint.

\subsection{Data acquisition and AOI design}
\begin{figure}[!htbp]
\centering
\includegraphics[width=0.88\linewidth]{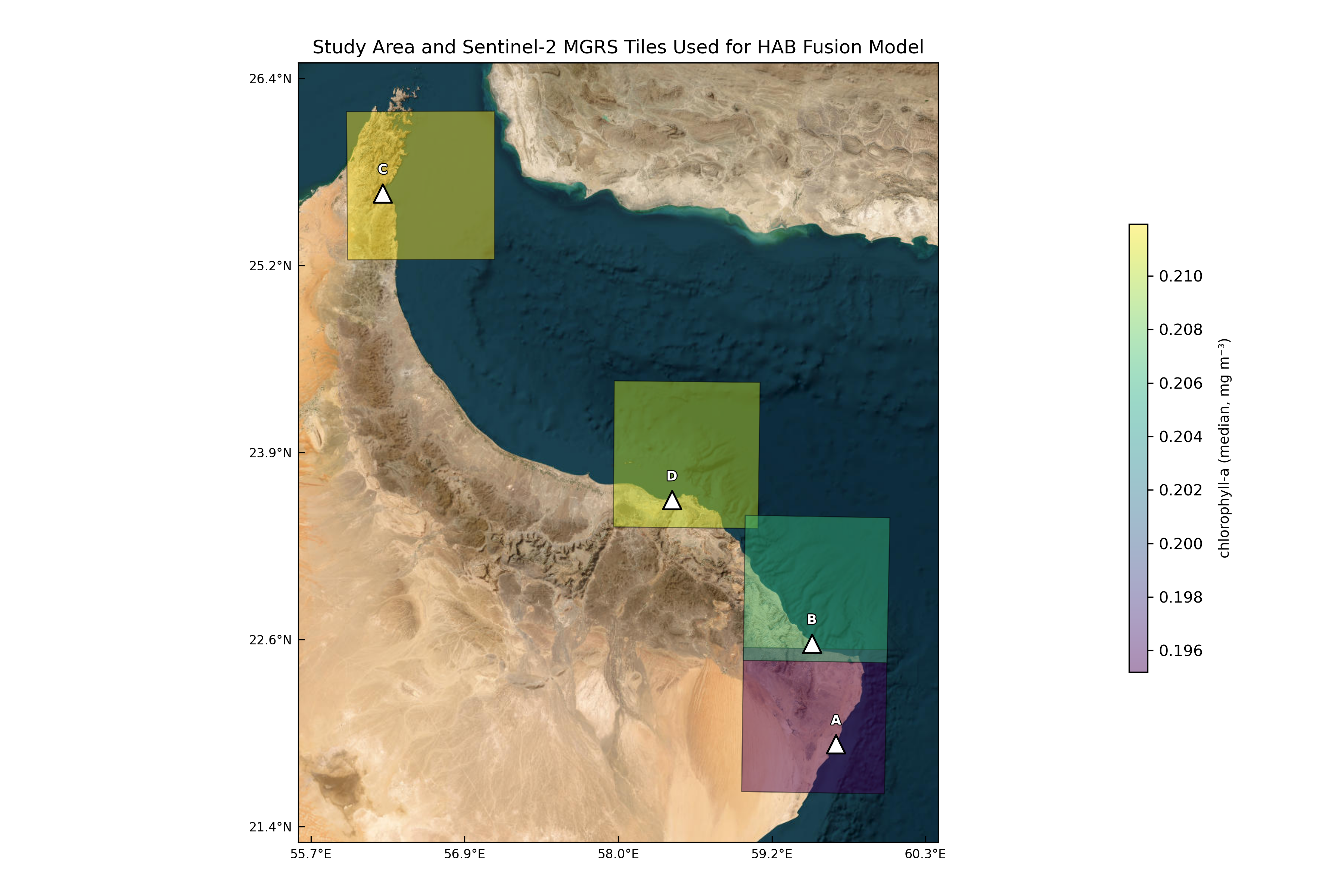}
\caption{Study area and Sentinel-2 MGRS tiles used for the HAB fusion model, with plant AOIs (A--D) shown for plant-centric chipping and aggregation.}
\label{fig:study_area}
\end{figure}

AOIs are defined around a set of coastal plants along Oman’s shoreline and buffered to capture nearshore waters relevant to intake risk. Sentinel-2 scenes are accessed via STAC search and retrieval (Planetary Computer), filtering by cloud constraints and temporal ranges \citep{planetary_computer_stac,esa_s2_guide}. MODIS Level-3 ocean color and thermal products are downloaded and time-aligned to provide context predictors \citep{nasa_oc_modis_l3}. Figure~\ref{fig:study_area} provides the study context and Sentinel-2 tile coverage used for plant-centric chipping and fusion.

\begin{figure}[!htbp]
\centering
\includegraphics[width=0.98\linewidth]{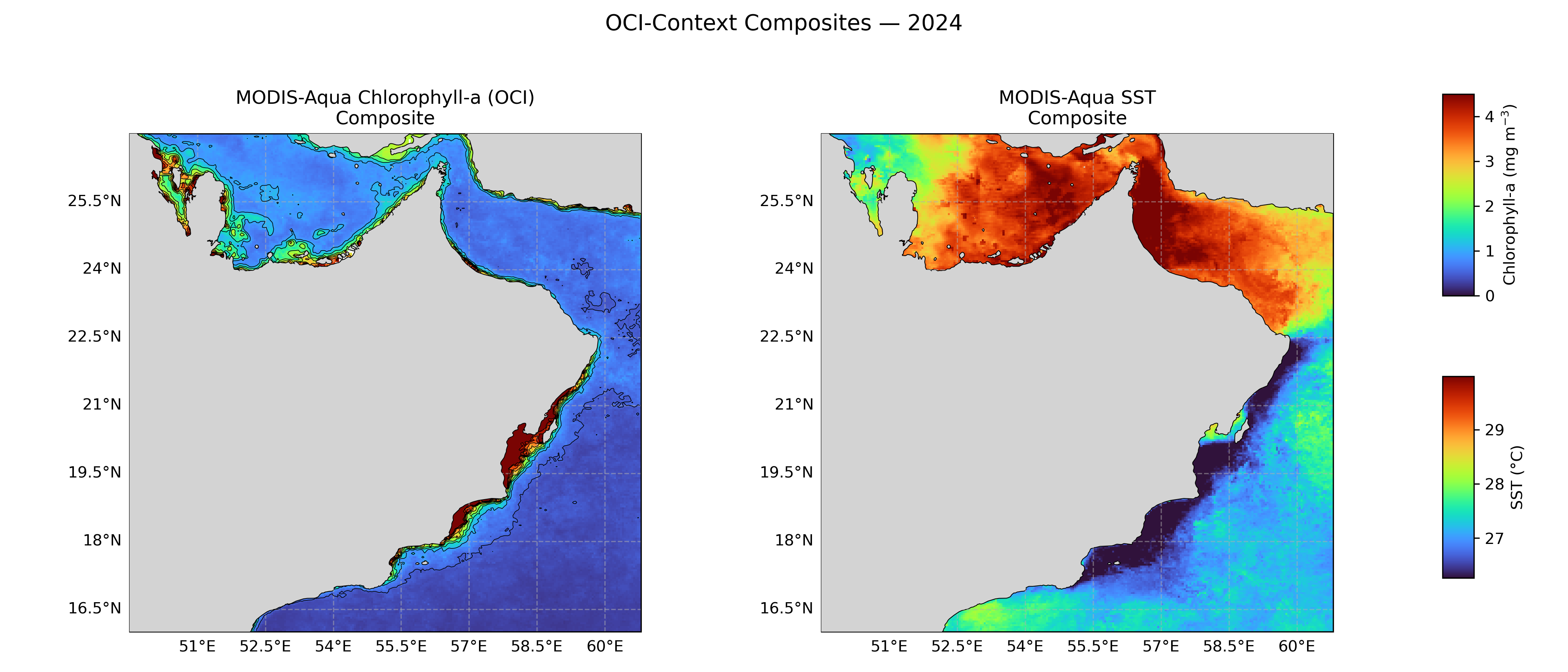}
\caption{OCI-context composites used by REDNET-ML: MODIS-Aqua chlorophyll-\textit{a} (left) and SST (right) over the Oman domain (2024 aggregate).}
\label{fig:oci_modis_chl_sst_2024}
\end{figure}
\FloatBarrier

\subsection{Sentinel-2 chipping and feature extraction}
For each AOI-date, the pipeline extracts fixed-size chips around the plant vicinity and computes summary statistics over valid water pixels. Features include indices such as NDWI \citep{mcf_ndwi_1996} and FAI-inspired anomalies \citep{hu_fai_2009} (plus red/NIR contrast ratios), robust summaries (mean/SD/quantiles) with invalid-pixel masking to reduce cloud/land contamination, and explicit seasonality encoding $(\sin(2\pi m/12), \cos(2\pi m/12))$.

\subsection{Label strategy: trusted labels and label mining}
Because HAB labels are sparse, two label views are maintained: trusted labels (conservative subset from reliable sources/manual checks) and weak/mined labels (heuristic candidates filtered to reduce false positives). The final label column is a controlled merge, enabling broad coverage while retaining trusted-only evaluation.

\begin{algorithm}[!htbp]
\caption{Sketch of label mining and non-leaky dataset construction}
\begin{algorithmic}[1]
\Require Time-indexed samples $(i)$; group key $g(i)$ (scene/overpass); trusted labels $y^{\text{trusted}}_i$ (possibly missing)
\Ensure Final labels $y^{\text{final}}_i$ and non-leaky splits
\State Compute heuristic score $h_i$ (e.g., thresholded ocean color/thermal anomalies)
\State Candidate positives $\mathcal{C} \leftarrow \{i: h_i = 1 \ \wedge\ \text{quality}(i)\text{ passes}\}$
\State Define weak labels $y^{\text{weak}}_i \leftarrow \mathbb{I}[i \in \mathcal{C}]$
\State Final label $y^{\text{final}}_i \leftarrow y^{\text{trusted}}_i \ \lor\ y^{\text{weak}}_i$
\State Split data by (a) grouping key $g(i)$ for group-safe CV or (b) time ranges for temporal CV
\end{algorithmic}
\end{algorithm}

\subsection{Decision fusion model and operating policy}
CatBoost is used as the decision fusion model due to strong tabular performance and stable handling of non-linear interactions \citep{prokhorenkova_catboost_2018}. Inputs include detector score summaries (e.g., max/median confidence, counts), Sentinel-derived index summaries, MODIS predictors (chlorophyll proxies, SST), and a small set of interaction features (e.g., $\log(\text{chlor\_a})$, $\text{chlor\_a}/\text{kd490}$). Thresholds are selected to satisfy operational monitoring needs: achieve minimum recall (e.g., $0.60$) while maximizing precision, aligning with PR behavior under imbalance \citep{davis_pr_roc_2006}.

In the current viewer build, the operational index is implemented as \texttt{ops\_risk\_v2\_seasonal} and is explicitly defined as:
\begin{align}
\tilde{x}_{j,t} &= \text{clip}\!\left(\frac{x_{j,t}-Q_{0.1}(x_j)}{Q_{0.9}(x_j)-Q_{0.1}(x_j)},\,0,\,1\right),\quad
j \in \{\text{chlor\_a},\text{nflh},\text{kd490},\text{sst}\} \\
\text{OCI}_t &= \frac{\sum_j w_j \tilde{x}_{j,t}\,\mathbf{1}_{j,t}}{\sum_j w_j\,\mathbf{1}_{j,t}}, \qquad
\mathbf{w}=(0.35,\,0.35,\,0.20,\,0.10) \\
c_t &= \frac{\sum_j w_j\,\mathbf{1}_{j,t}}{\sum_j w_j}, \qquad
\text{OCI}^{\text{adj}}_t=\text{OCI}_t\,c_t + 0.5(1-c_t),
\end{align}
where $\mathbf{1}_{j,t}$ indicates feature availability. In this pipeline, zero values in \texttt{chlor\_a}, \texttt{kd490}, and \texttt{nflh} are treated as missing placeholders before normalization.

Seasonality is incorporated per driver with monthly anomaly z-scores:
\begin{align}
z_{j,t} &= \text{clip}\!\left(\frac{x_{j,t}-\text{median}_{m(t)}(x_j)}{\text{IQR}_{m(t)}(x_j)},\,-3,\,3\right), \\
s_{j,t} &= \sigma(1.15\,z_{j,t}),
\end{align}
(with SST anomaly clipped at zero before the sigmoid), then aggregated with the same driver ratios to produce \(\text{Season}^{\text{adj}}_t\).

The final operational blend is:
\begin{align}
b_t &= 0.40\,\text{hab\_prob}_t + 0.25\,\text{det\_mean}_t + 0.20\,\text{OCI}^{\text{adj}}_t + 0.15\,\text{Season}^{\text{adj}}_t, \\
d_t &= \text{clip}\!\left(1 - 0.18\left|\text{hab\_prob}_t-\text{det\_mean}_t\right|,\,0.75,\,1.0\right), \\
\text{ops\_risk}_t &= \text{clip}(b_t d_t,\,0,\,1),
\end{align}
with fallback to \(\text{hab\_prob}_t\) if \(\text{ops\_risk}_t\) is missing.

Thresholds are calibrated by matching the exceedance rates of legacy HAB thresholds:
\[
r_{\text{watch}}=\Pr(\text{hab\_prob}\ge 0.55), \quad
r_{\text{action}}=\Pr(\text{hab\_prob}\ge 0.6238688594),
\]
\[
\tau_{\text{watch}} = Q_{1-r_{\text{watch}}}(\text{ops\_risk}), \quad
\tau_{\text{action}} = Q_{1-r_{\text{action}}}(\text{ops\_risk}),
\]
with constraints \(\tau_{\text{action}}-\tau_{\text{watch}}\ge 0.04\), \(\tau_{\text{watch}}\in[0.05,0.95]\), and \(\tau_{\text{action}}\in[0.05,0.99]\).

\begin{table}[!htbp]
\centering
\small
\caption{Decision layer summary: model choices and how they are used at inference time.}
\label{tab:model_decisions}
\begin{tabular}{@{}p{3.0cm}p{11.5cm}@{}}
\toprule
Decision & Implemented choice and rationale \\ \midrule
Image evidence & Detectors provide score summaries (max/median/count) rather than final labels to improve robustness under sparse/noisy labels. \\
Detectors kept & Faster R-CNN (ResNet50) and SSD MobileNet retained due to strong region-level performance. \\
Detector excluded & YOLOv8 evaluated but excluded to reduce complexity (no material ensemble gain) \citep{ultralytics_yolov8}. \\
Fusion model & CatBoost decision fusion on MODIS, indices and detector scores \citep{prokhorenkova_catboost_2018}. \\
Operating metrics & Emphasis on AUPRC (primary) and AUROC (support) due to imbalance \citep{davis_pr_roc_2006}. \\
Alerting policy & Two thresholds: $\tau_{\text{watch}}$ (candidate surfacing) and $\tau_{\text{action}}$ (higher-confidence escalation). \\
Shift awareness & Score drift tracked (PSI and KS) to contextualize threshold transferability. \\ \bottomrule
\end{tabular}
\end{table}

The artifact script also exports a feature-range summary. Table~\ref{tab:ranges_overall} documents the numerical scale of the key predictors used by the fusion model.

\begin{table}[!htbp]
\centering
\small
\caption{Range of data used (training table): numerical ranges and mean $\pm$ SD for key predictors.}
\label{tab:ranges_overall}
\begin{tabular}{@{}lcc@{}}
\toprule
Parameter & Range & Mean $\pm$ SD \\ \midrule
\texttt{chlor\_a}     & 0.044--9.972   & 0.312 $\pm$ 0.534 \\
\texttt{kd490}        & 0.019--2.244   & 0.056 $\pm$ 0.072 \\
\texttt{nflh}         & -0.026--0.736  & 0.124 $\pm$ 0.109 \\
\texttt{sst}          & 23.945--32.995 & 28.008 $\pm$ 1.853 \\
\texttt{fai\_mean}    & -0.073--0.084  & 0.028 $\pm$ 0.025 \\
\texttt{ndwi\_mean}   & -0.393--0.068  & -0.156 $\pm$ 0.074 \\
\texttt{rednir\_mean} & 0.367--1.088   & 0.889 $\pm$ 0.058 \\
\bottomrule
\end{tabular}
\end{table}
\FloatBarrier

\section{Implementation Details}

\subsection{Repository organization and reproducibility}
REDNET-ML is implemented as a modular repository with notebook-first development and script equivalents for batch execution. Notebooks serve as an executable ``lab book'' (AOI setup, data acquisition, chipping/indices, label mining, detector training, fusion training, temporal benchmarking, inference, and viewer export), while scripts generate reproducible run artifacts (metrics, plots, and serialized models). The report artifact workflow exports the ROC/PR curves, confusion matrices, calibration plots, feature importance summaries, and SHAP explanations into a LaTeX-ready directory. As of February 27, 2026, the project test suite includes 53 passing \texttt{pytest} checks across viewer logic, HAB preparation/scoring, download helpers, evaluation metrics, and utility modules.

\subsection{Architecture overview}
The pipeline is organized into: (i) a \emph{Data Layer} (raw satellite data, label tables, chip tables), (ii) a \emph{Detection Layer} (Torchvision/YOLO detectors), (iii) an \emph{Inference \& Postprocessing Layer} (chip scoring, feature assembly, plant-time aggregation), (iv) a \emph{Geospatial Analysis Layer} (centroids, spatial joins, plant proximity), and (v) a \emph{Reporting \& Visualization Layer} (artifact generation and risk-field viewer). This structure allows swapping detectors or fusion models without breaking end-to-end inference.

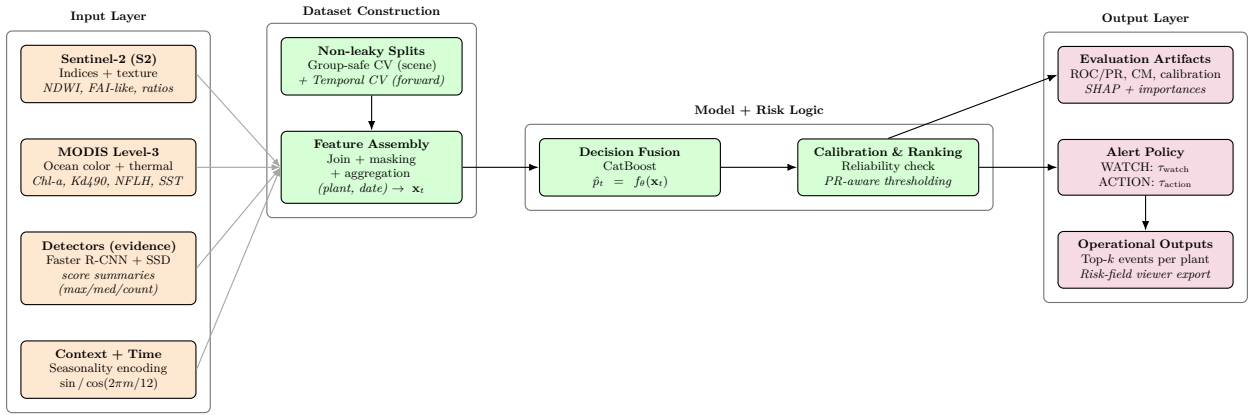
\begin{figure}[H]
\centering
\resizebox{\linewidth}{!}{%
\begin{tikzpicture}[
font=\small,
node distance=0.9cm and 1.35cm,
box/.style={draw,rounded corners,thick,align=center,inner sep=6pt,minimum height=10mm},
inbox/.style={box,fill=orange!18,minimum width=4.05cm,text width=4.05cm},
midbox/.style={box,fill=green!16,minimum width=4.2cm,text width=4.2cm},
outbox/.style={box,fill=purple!14,minimum width=4.05cm,text width=4.05cm},
arrow/.style={-{Latex[length=2.5mm]},thick},
faint/.style={draw=black!35,thick,-{Latex[length=2.3mm]}}
]

\node[inbox] (s2) {\textbf{Sentinel-2 (S2)}\\Indices + texture\\\emph{NDWI, FAI-like, ratios}};
\node[inbox,below=of s2] (modis) {\textbf{MODIS Level-3}\\Ocean color + thermal\\\emph{Chl-a, Kd490, NFLH, SST}};
\node[inbox,below=of modis] (det) {\textbf{Detectors (evidence)}\\Faster R-CNN + SSD\\\emph{score summaries (max/med/count)}};
\node[inbox,below=of det] (time) {\textbf{Context + Time}\\Seasonality encoding\\\emph{$\sin/\cos(2\pi m/12)$}};

\node[midbox,right=2.15cm of modis] (assemble) {\textbf{Feature Assembly}\\Join + masking + aggregation\\\emph{(plant, date) $\rightarrow \mathbf{x}_t$}};
\node[midbox,above=of assemble] (splits) {\textbf{Non-leaky Splits}\\Group-safe CV (scene)\\\emph{+ Temporal CV (forward)}};
\node[midbox,right=1.95cm of assemble] (cat) {\textbf{Decision Fusion}\\CatBoost\\\emph{$\hat{p}_t = f_\theta(\mathbf{x}_t)$}};
\node[midbox,right=1.95cm of cat] (cal) {\textbf{Calibration \& Ranking}\\Reliability check\\\emph{PR-aware thresholding}};

\node[outbox,right=2cm of cal] (policy) {\textbf{Alert Policy}\\WATCH: $\tau_{\text{watch}}$\\ACTION: $\tau_{\text{action}}$};
\node[outbox,above=of policy] (eval) {\textbf{Evaluation Artifacts}\\ROC/PR, CM, calibration\\\emph{SHAP + importances}};
\node[outbox,below=of policy] (viewer) {\textbf{Operational Outputs}\\Top-$k$ events per plant\\\emph{Risk-field viewer export}};

\draw[faint] (s2.east) -- (assemble.west);
\draw[faint] (modis.east) -- (assemble.west);
\draw[faint] (det.east) -- (assemble.west);
\draw[faint] (time.east) -- (assemble.west);
\draw[arrow] (splits.south) -- (assemble.north);
\draw[arrow] (assemble) -- (cat);
\draw[arrow] (cat) -- (cal);
\draw[arrow] (cal) -- (policy);
\draw[arrow] (cal.north) -- (eval.west);
\draw[arrow] (policy.south) -- (viewer.north);

\begin{scope}[on background layer]
\node[draw=black!55,rounded corners,thick,inner sep=10pt,fit=(s2)(time)] (inputlayer) {};

\node[draw=black!55,rounded corners,thick,inner sep=10pt,
      fit=(splits)(assemble)] (datasetlayer) {};

\node[draw=black!55,rounded corners,thick,inner sep=10pt,
      fit=(cat)(cal)] (modellayer) {};

\node[draw=black!55,rounded corners,thick,inner sep=10pt,fit=(eval)(viewer)] (outlayer) {};
\end{scope}

\node[above=2pt of inputlayer.north] {\textbf{Input Layer}};
\node[above=2pt of datasetlayer.north] {\textbf{Dataset Construction}};
\node[above=2pt of modellayer.north] {\textbf{Model + Risk Logic}};
\node[above=2pt of outlayer.north] {\textbf{Output Layer}};

\end{tikzpicture}%
}
\caption{REDNET-ML architecture: multi-sensor evidence generation, non-leaky dataset construction, CatBoost decision fusion, and operational risk outputs.}
\label{fig:rednet_arch}
\end{figure}
\FloatBarrier

\subsection{Non-leaky identifiers and dataset construction}
To prevent leakage, each chip is associated with a stable tile/chip ID (scene \& location) and a scene group key so that chips from the same overpass are never split across folds. A plant ID is used for per-site aggregation and for deployment-ready partitioning. The training table is built by joining Sentinel summaries, time-aligned MODIS predictors, detector score summaries, and trusted/weak/final labels.

\subsection{Detector training and score extraction}
Detectors are trained on COCO-formatted datasets derived from positive-labeled chips. Torchvision (Faster R-CNN, SSD) enables controlled, reproducible configurations \citep{ren_fasterrcnn_2015,liu_ssd_2016}; YOLO provides a strong one-stage baseline \citep{ultralytics_yolov8}. At inference, each detector outputs bounding boxes with confidences; these are aggregated into compact tabular features (e.g., max confidence, median confidence of filtered detections, counts above a threshold) and normalized for fusion training stability.

\subsection{Fusion training and artifact generation}
Fusion training is implemented to produce per-fold predictions and evaluation artifacts (AUROC/AUPRC, ROC/PR curves, confusion matrices at selected thresholds, calibration curves, and feature importance summaries). A dedicated artifact script consolidates plots and exports them for direct LaTeX inclusion.

\subsection{Deployment: end-to-end inference and risk viewer}
The deployment pipeline consumes the trained fusion model and generates per-plant outputs with $(t, \hat{p}_{t}, \text{state})$, alongside supporting predictors and detector summaries. Outputs support operational review (ranking top-$k$ risk events) and risk-field generation, which converts high-risk events into spatial overlays for interactive exploration in a web viewer.

\begin{figure}[!ht]
\centering
\includegraphics[width=\linewidth]{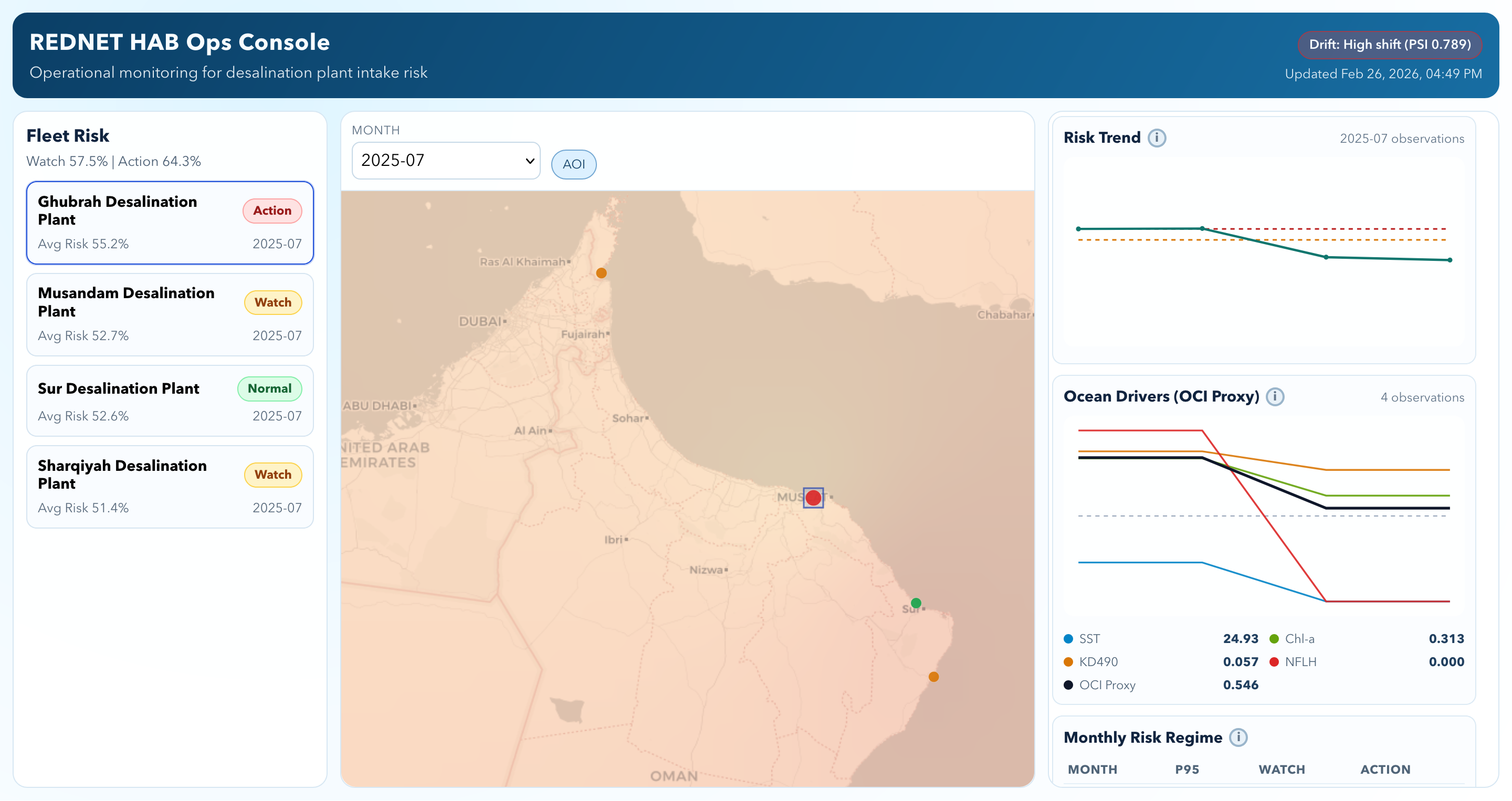}
\caption{REDNET HAB Ops Console (full GUI) used for operational exploration of plant risk, monthly context, and event-level evidence.}
\label{fig:viewer_gui_full}
\end{figure}
\FloatBarrier

\section{Results, Evaluation, and Conclusions}

\subsection{Evaluation protocol and metrics}
Evaluation targets (i) discriminative ability (AUROC/AUPRC) and (ii) operational usefulness (precision/recall tradeoffs under threshold policies). Because HAB data are imbalanced and the monitoring objective prioritizes surfacing candidates, AUPRC is treated as the primary metric with AUROC as supporting context \citep{davis_pr_roc_2006}.

\subsection{Cross-validation results}
Under group-safe cross-validation (scene-aware folds), the all-label fusion model achieves mean AUROC $= 0.842 \pm 0.019$ and mean AUPRC $= 0.731 \pm 0.029$ across 5 folds. Pooling predictions across folds yields pooled AUROC $= 0.842$ and pooled AUPRC $= 0.727$ at an operating threshold of $\approx 0.454$ chosen under a minimum recall constraint ($\ge 0.60$). At this point, pooled confusion counts are $(TP, FP, TN, FN) = (398, 185, 887, 260)$, corresponding to precision $\approx 0.682$ and recall $\approx 0.605$. The positive-class prevalence in the pooled evaluation set is $\approx 0.268$, matching the PR ``no-skill'' baseline.

\begin{figure}[!ht]
\centering
\begin{subfigure}[b]{0.49\linewidth}
\centering
\includegraphics[width=\linewidth]{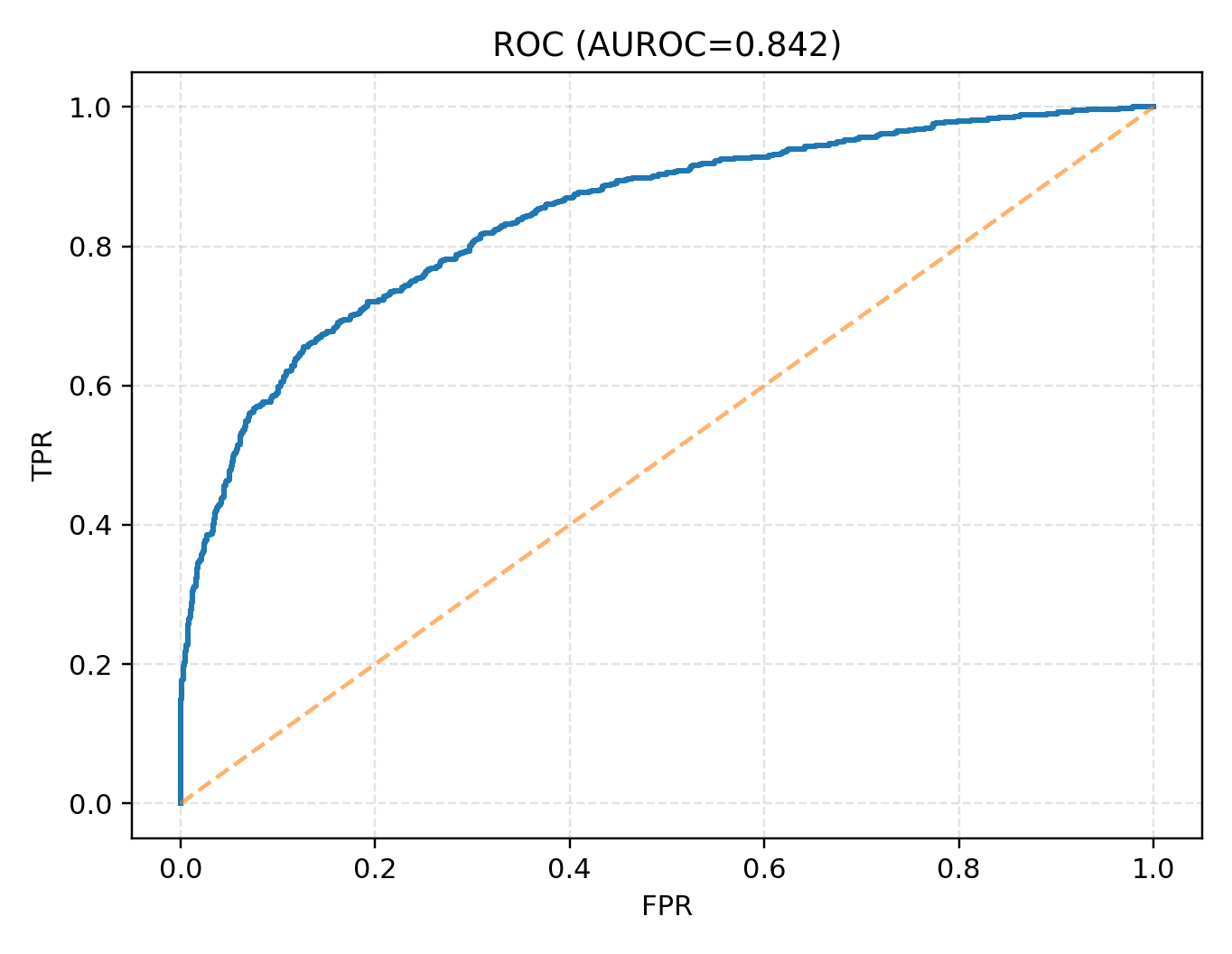}
\caption{Pooled ROC curve.}
\label{fig:roc_pooled}
\end{subfigure}
\hfill
\begin{subfigure}[b]{0.49\linewidth}
\centering
\includegraphics[width=\linewidth]{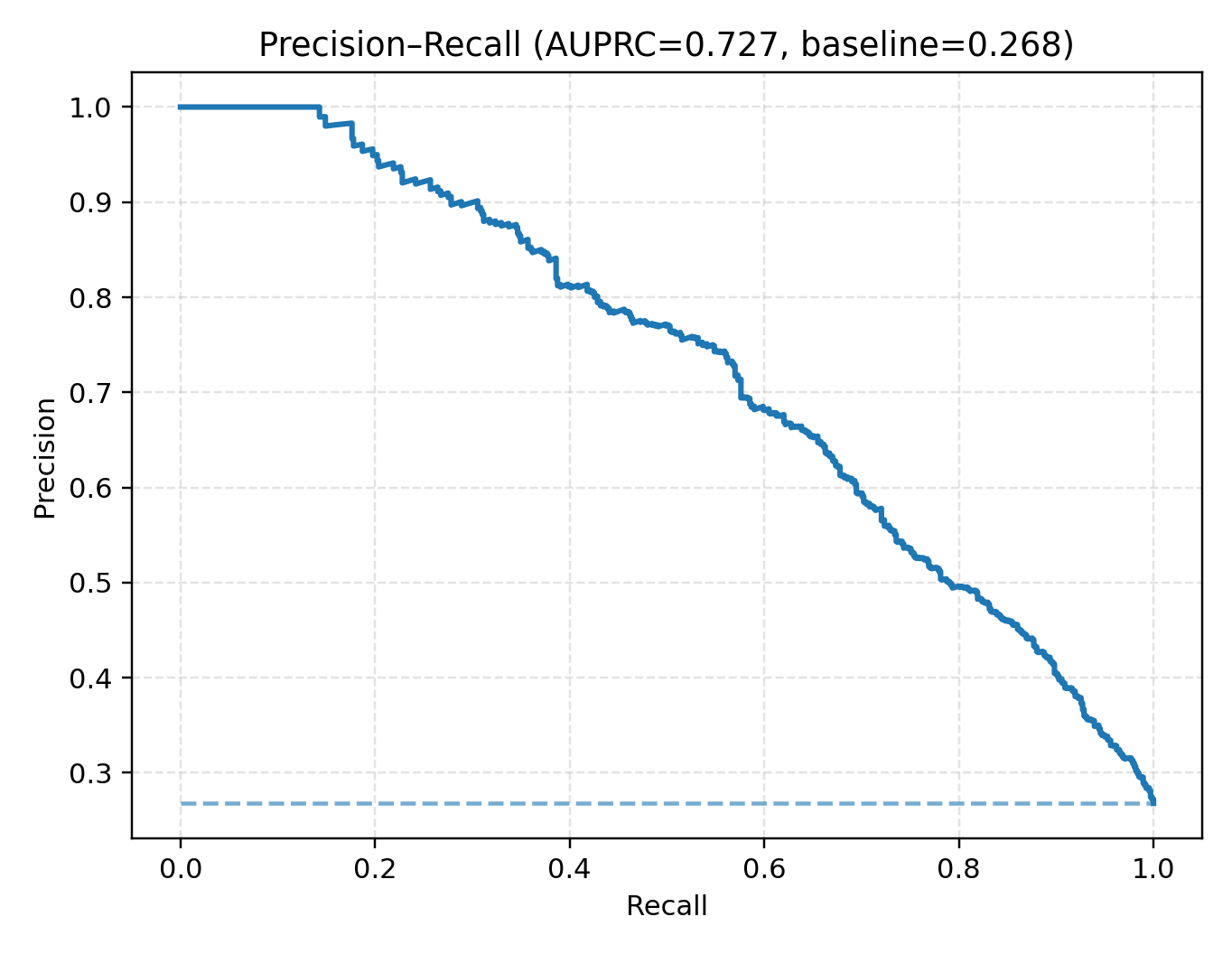}
\caption{Pooled PR curve.}
\label{fig:pr_pooled}
\end{subfigure}
\caption{Pooled ROC and PR curves.}
\label{fig:rocpr_pooled}
\end{figure}
\FloatBarrier

\subsection{Detector benchmarking and ensemble decision}
Detectors are compared using region-level metrics rather than standard mAP. Since HABs exhibit weak boundaries and spatial ambiguity, evaluation emphasizes regional recall and mean area overlap as monitoring-relevant measures.

\begin{table}[!ht]
\centering
\small
\caption{Detector comparison using region-level metrics ($n=51$ ground-truth regions).}
\label{tab:detector_bench}
\begin{tabular}{@{}lccp{5.2cm}@{}}
\toprule
Model & Regional Recall & Mean Area Overlap & Takeaway \\ \midrule
Faster R-CNN ResNet50 & 0.94 & 0.69 & Best overall accuracy and localization \\
SSD MobileNet & 0.94 & 0.66 & Best efficiency--accuracy tradeoff \\
Faster R-CNN MobileNet & 0.44 & 0.31 & Capacity-limited; light baseline only \\
YOLOv8n3 & 0.92 & 0.50 & Competitive recall; weaker alignment for this task \\
\bottomrule
\end{tabular}
\end{table}

The final system uses Faster R-CNN (ResNet50) and SSD MobileNet as complementary evidence generators; YOLOv8 did not materially increase ensemble coverage and was excluded to reduce complexity.
\FloatBarrier

\subsection{Calibration and thresholding (WATCH vs ACTION)}
Calibration is evaluated via reliability curves; it is treated cautiously because limited calibration positives in some folds can destabilize Platt scaling or isotonic regression \citep{platt_1999,zadrozny_isotonic_2002}. In the deployed viewer logic, base thresholds are fixed at \(\tau_{\text{watch,base}}=0.55\) and \(\tau_{\text{action,base}}=0.6238688594\), then operational \(\tau_{\text{watch}},\tau_{\text{action}}\) are re-fitted by quantile matching to legacy alert-load rates on the current calibration pool, with a minimum enforced gap of \(0.04\). Operationally, \(\hat{p}_t\) is used primarily for ranking and then thresholded into a two-state policy: \textsf{WATCH} for review (high sensitivity) and \textsf{ACTION} for escalation (higher confidence). Under missing-label fallback conditions, the static base thresholds are retained.

\begin{figure}[!ht]
\centering
\begin{subfigure}[b]{0.49\linewidth}
\centering
\includegraphics[width=\linewidth]{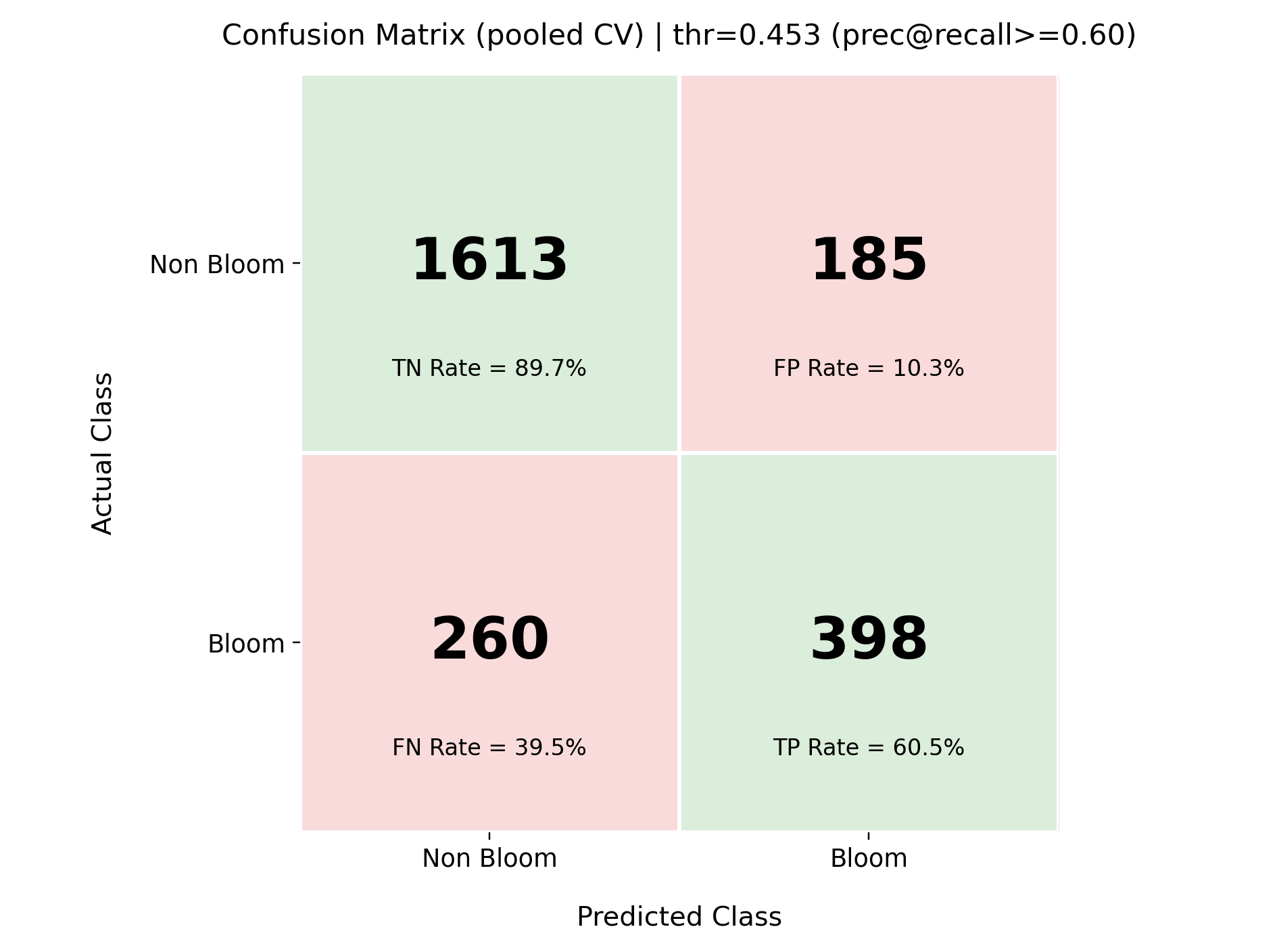}
\caption{Confusion matrix at selected threshold.}
\label{fig:cm}
\end{subfigure}
\hfill
\begin{subfigure}[b]{0.49\linewidth}
\centering
\includegraphics[width=\linewidth]{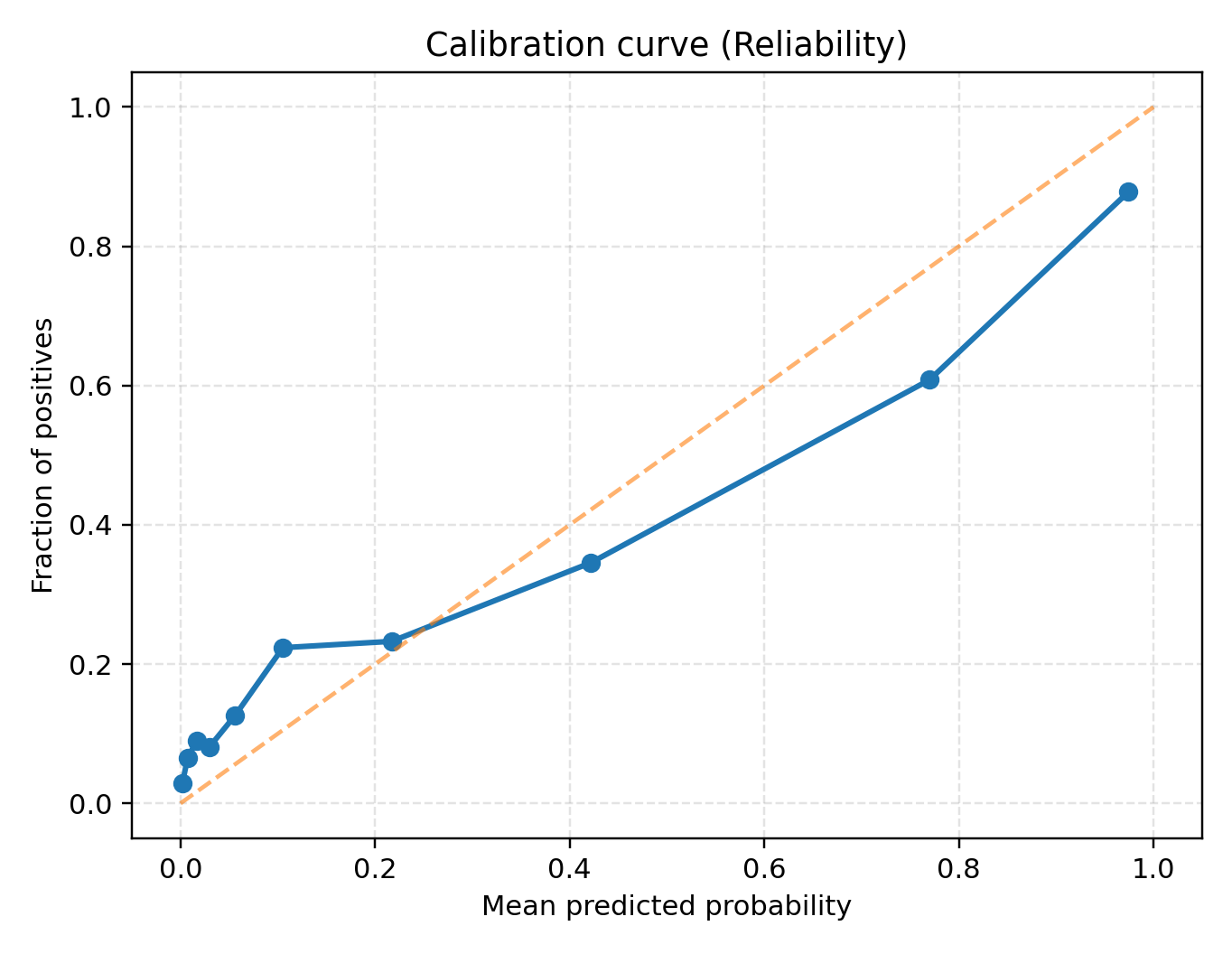}
\caption{Calibration / reliability curve.}
\label{fig:calib}
\end{subfigure}
\caption{Thresholded and calibration artifacts for the fusion model.}
\label{fig:threshold_calibration_artifacts}
\end{figure}
\FloatBarrier

\subsection{Drift and generalization to 2025}
Generalization is tested by comparing 2017--2024 score distributions against 2025 per plant using PSI and KS drift metrics, evaluating monthly alert rates under the thresholds above, and exporting top-$k$ events per plant for inspection. A representative distribution check indicates shape shift (lower central tendency but heavier upper tail), consistent with high KS distance and high PSI. Observed per-plant PSI ranges from 1.4 to 5.2 and KS distance ranges from 0.44 to 0.67, while pooled PSI can be lower (e.g., $\sim 0.789$) due to smoothing across plants. The implication is that thresholds selected on older validation distributions may not transfer cleanly; \textsf{WATCH} is therefore maintained as a robust candidate-surfacing threshold, while \textsf{ACTION} may require periodic recalibration under drift.

\subsection{Compact artifact summary}
To keep the report concise while preserving evidence, the key exported artifacts are placed immediately after the relevant paragraphs above: ROC/PR after cross-validation, confusion/calibration after thresholding, and interpretability below. These figures are produced by the repository artifact scripts and included directly from the LaTeX artifact folders.

\begin{figure}[!ht]
\centering
\begin{subfigure}[b]{0.49\linewidth}
\centering
\includegraphics[width=\linewidth]{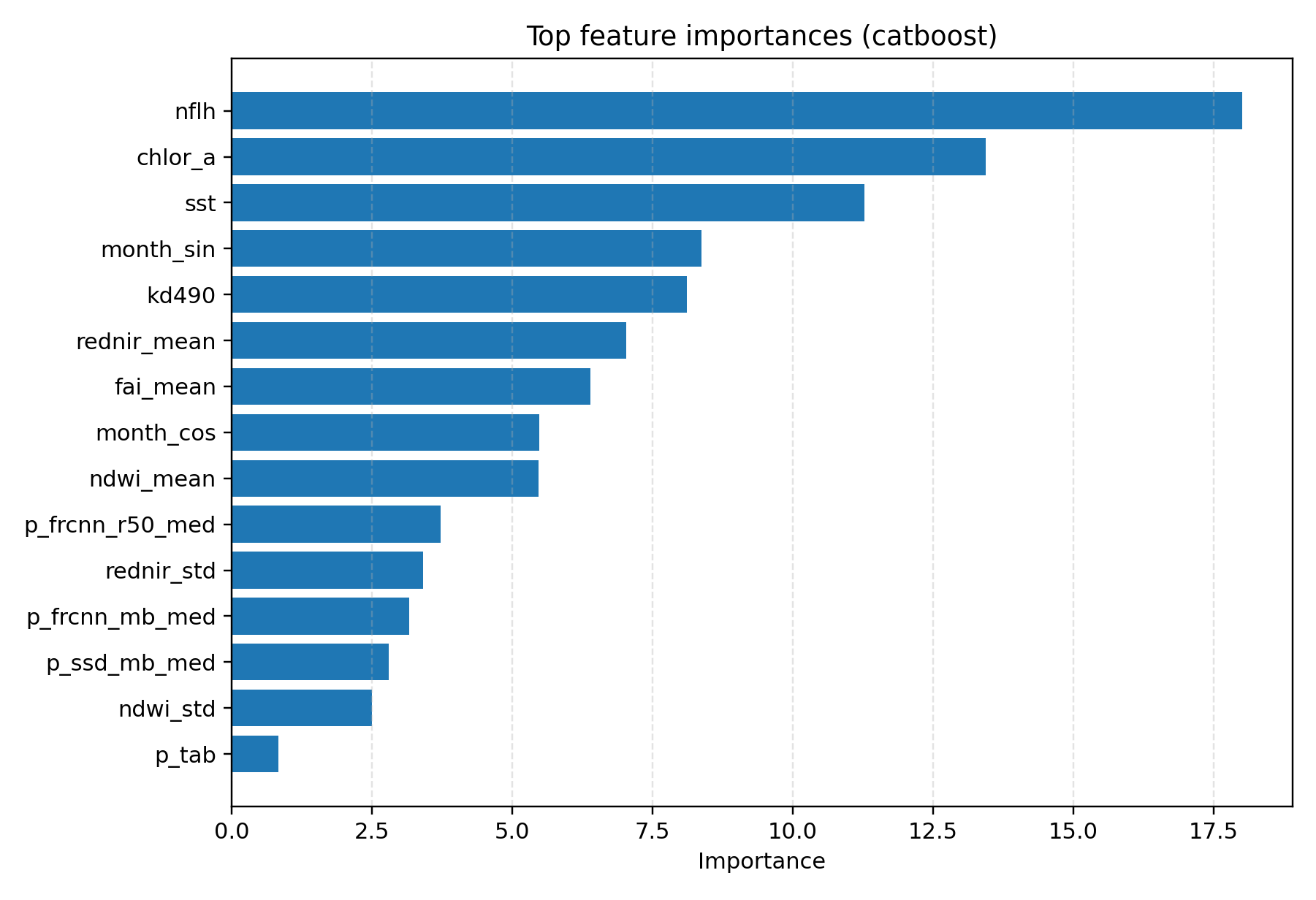}
\caption{Top feature importances (CatBoost).}
\label{fig:featimp}
\end{subfigure}
\hfill
\begin{subfigure}[b]{0.49\linewidth}
\centering
\includegraphics[width=\linewidth]{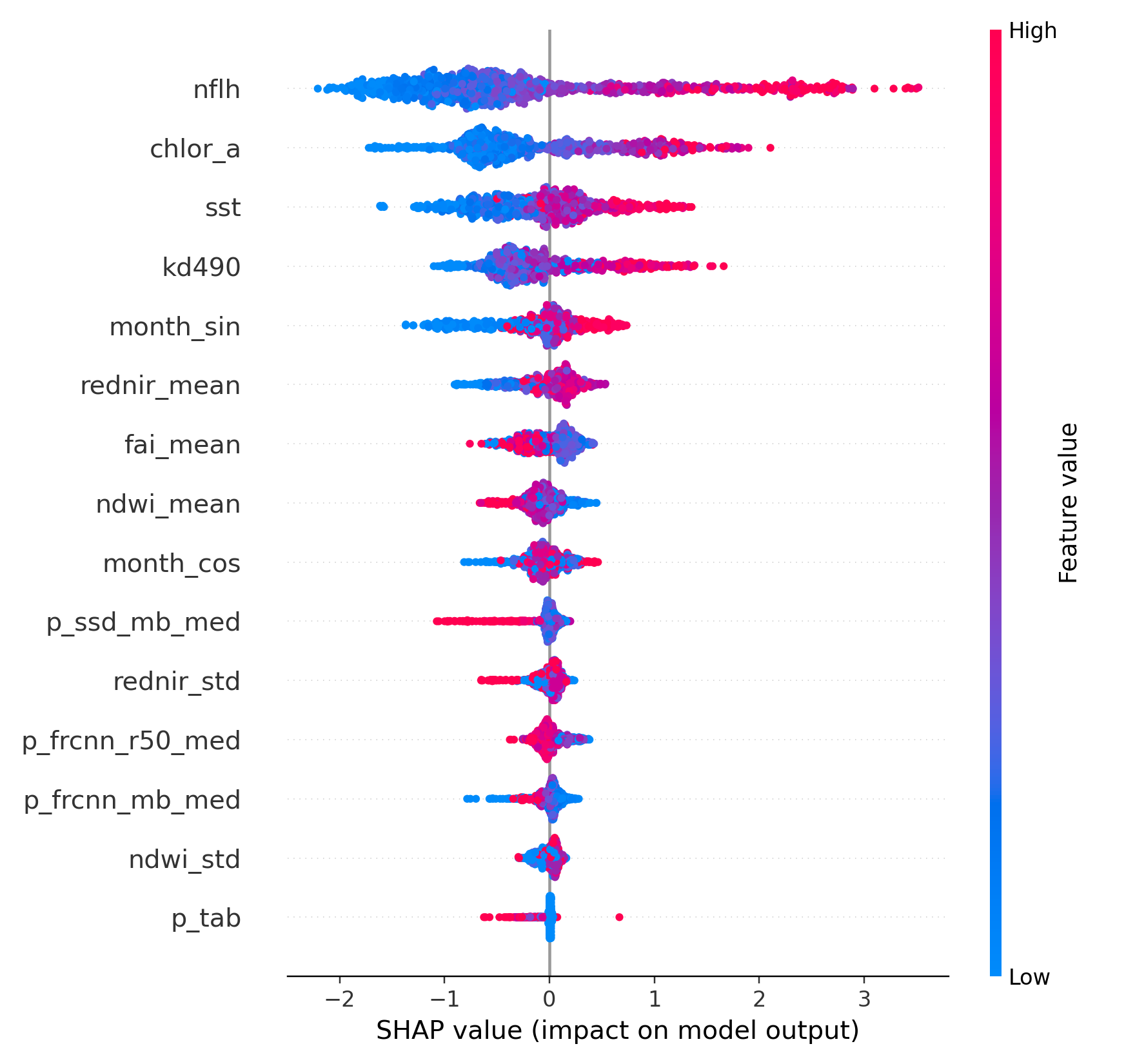}
\caption{SHAP summary (global feature effects).}
\label{fig:shap_summary}
\end{subfigure}
\caption{Model interpretability artifacts exported by the report scripts.}
\label{fig:featimp_shap}
\end{figure}
\FloatBarrier

\subsection{Conclusions}
REDNET-ML demonstrates that a carefully engineered, multi-sensor fusion approach can produce an operationally meaningful HAB risk signal under realistic constraints: noisy labels, limited positives, and strong temporal dependence. The project’s main strengths are its non-leaky evaluation design, modular implementation, and inspection-ready artifacts that make the system transparent and reproducible.

Limitations are primarily tied to label sparsity and temporal distribution shift. Drift metrics in 2025 indicate that thresholds may not remain stable over time; this supports using a robust \textsf{WATCH} policy for candidate surfacing and treating \textsf{ACTION} as a higher-confidence decision that may require periodic recalibration or retraining as distributions shift.

\bibliographystyle{unsrtnat}
\bibliography{references}

\end{document}